%% file: 00_main.tex
\documentclass[conference]{IEEEtran}
\IEEEoverridecommandlockouts
\usepackage{cite}
\usepackage{amsmath,amssymb,amsfonts}
\usepackage{algorithmic}
\usepackage{graphicx}
\usepackage{textcomp}
\usepackage{xcolor}
\usepackage{url}
\def\BibTeX{{\rm B\kern-.05em{\sc i\kern-.025em b}\kern-.08em
    T\kern-.1667em\lower.7ex\hbox{E}\kern-.125emX}}
\begin{document}

\title{
    Real-Time Fish Detection in Indonesian Marine Ecosystems Using Lightweight YOLOv10-Nano Architecture\\
    \thanks{
    Preprint. This manuscript has been accepted for publication in \textit{Jurnal Teknik Informatika}. The final version will be available in 2026.
    }
}

\author{
Jonathan Wuntu, Dwisnanto Putro, Rendy Syahputra\\
Dept. of Electrical Engineering\\ Sam Ratulangi University}

\maketitle

\input{01_abstract}
\input{02_intro}
\input{03_method}
\input{04_experiment}
\input{05_conclusion}

\section*{Acknowledgment}
Special thanks and appreciation to everyone in the AIVision research group whose efforts and support were instrumental in bringing this work to completion.

\bibliographystyle{ieeetr} % This style formats the bibliography
\bibliography{egbib} % This links to your egbib.bib file

\end{document}

%% file: 01_abstract.tex
\begin{abstract}
Indonesia's marine ecosystems, part of the globally recognized Coral Triangle, are among the richest in biodiversity, requiring efficient monitoring tools to support conservation. Traditional fish detection methods are time-consuming and demand expert knowledge, prompting the need for automated solutions. This study explores the implementation of YOLOv10-nano, a state-of-the-art deep learning model, for real-time marine fish detection in Indonesian waters, using test data from Bunaken National Marine Park. YOLOv10’s architecture, featuring improvements like the CSPNet backbone, PAN for feature fusion, and Pyramid Spatial Attention Block, enables efficient and accurate object detection even in complex environments. The model was evaluated on the DeepFish and OpenImages V7-Fish datasets. Results show that YOLOv10-nano achieves a high detection accuracy with mAP50 of 0.966 and mAP50:95 of 0.606 while maintaining low computational demand (2.7M parameters, 8.4 GFLOPs). It also delivered an average inference speed of 29.29 FPS on the CPU, making it suitable for real-time deployment. Although OpenImages V7-Fish alone provided lower accuracy, it complemented DeepFish in enhancing model robustness. Overall, this study demonstrates YOLOv10-nano’s potential for efficient, scalable marine fish monitoring and conservation applications in data-limited environments.
\end{abstract}

\begin{IEEEkeywords}
deep learning; fish detection; Indonesian waters; marine conservations; object detection; YOLOv10
\end{IEEEkeywords}

%% file: 02_intro.tex
\section{Introduction}
Indonesian waters, as part of the globally recognized Coral Triangle, constitute one of the world's most biodiverse marine ecosystems, encompassing thousands of fish species, coral reefs, and marine organisms across numerous protected areas and national parks \cite{Widayanti_2022}. The archipelagic geography of Indonesia's 17,000+ islands creates diverse marine habitats from shallow reefs to deep-sea environments, making these ecosystems critically important for both local communities dependent on marine resources and global biodiversity conservation. Effective conservation of these complex marine systems requires accurate data on fish populations and distribution patterns, which attracts researchers and visitors who contribute to marine ecological understanding and support sustainability efforts throughout Indonesian waters \cite{Widayanti_2022}.

In recent years, advancements in artificial intelligence, particularly in deep learning \cite{math10224189}, \cite{Naflah_Faulina_2024}, have opened new possibilities for automating fish detection. Convolutional Neural Networks (CNNs) have emerged as powerful tools in image analysis, capable of identifying objects in images with high accuracy by extracting relevant visual features through layered processing \cite{Alzubaidi2021}, \cite{jmse13010135}. One of the most effective object detection frameworks built on CNNs is YOLO (You Only Look Once), which is known for its ability to perform real-time object detection in a single processing step.

YOLO has seen numerous iterations, each improving upon its speed and accuracy. The latest version, YOLOv10, released in 2024, incorporates several architectural enhancements such as a refined CSPNet backbone, a Path Aggregation Network (PAN) \cite{miracle2020panet} for multi-scale feature fusion, and dual detection heads (One2One and One2Many) to improve performance in complex scenes \cite{wang2024yolov10realtimeendtoendobject}, \cite{make5040083}. Additionally, YOLOv10 is more computationally efficient compared to its predecessor, YOLOv8, using significantly fewer parameters and lower GFLOPs while achieving better detection results \cite{hussain2024yolov5yolov8yolov10goto}, \cite{alif2024yolov1yolov10comprehensivereview}, \cite{sapkota2025yoloadvancesgenesisdecadal}. These improvements are further complemented by the PSA (Pyramid Spatial Attention) Block, which enhances the model’s ability to detect objects of varying shapes and sizes.

Specialized adaptations of YOLO, such as HRA-YOLO,  YOLO-Fish and AquaYOLO, have demonstrated promising results in underwater environments, showing increased accuracy and robustness in marine fish detection \cite{ECRS2023-16315}, \cite{MUKSIT2022101847}, \cite{Vijayalakshmi2025}, \cite{isprs-archives-XLVI-3-W1-2022-301-2022}.

Given these technological advancements, this study aims to apply the YOLOv10 model for marine fish detection in Indonesian waters and test with fish inside Bunaken National Marine Park using the DeepFish dataset \cite{Saleh2020} and OpenImages V7-Fish \cite{Kuznetsova_2020}. The DeepFish dataset provides a diverse sample of fish across twenty habitats of mangroves. On the other hand, the OpenImages V7-Fish dataset is a part of the OpenImages V7 dataset with only the class of fish, to produce accurate and reliable data to support ongoing conservation efforts. Therefore, the main contributions of this work are as follows:
\begin{enumerate}
  \item Implementing YOLOv10 for the purpose of marine fish detection and counting.
  \item Evaluating YOLOv10 real-time performance on a benchmark dataset.
  \item Analyzing the capabilities of YOLOv10n in terms of performance and latencies for marine fish detection.
\end{enumerate}

%% file: 03_method.tex
\begin{figure*}
    \vspace{-0.05in}
    \centering
    \includegraphics[width=0.83\linewidth]{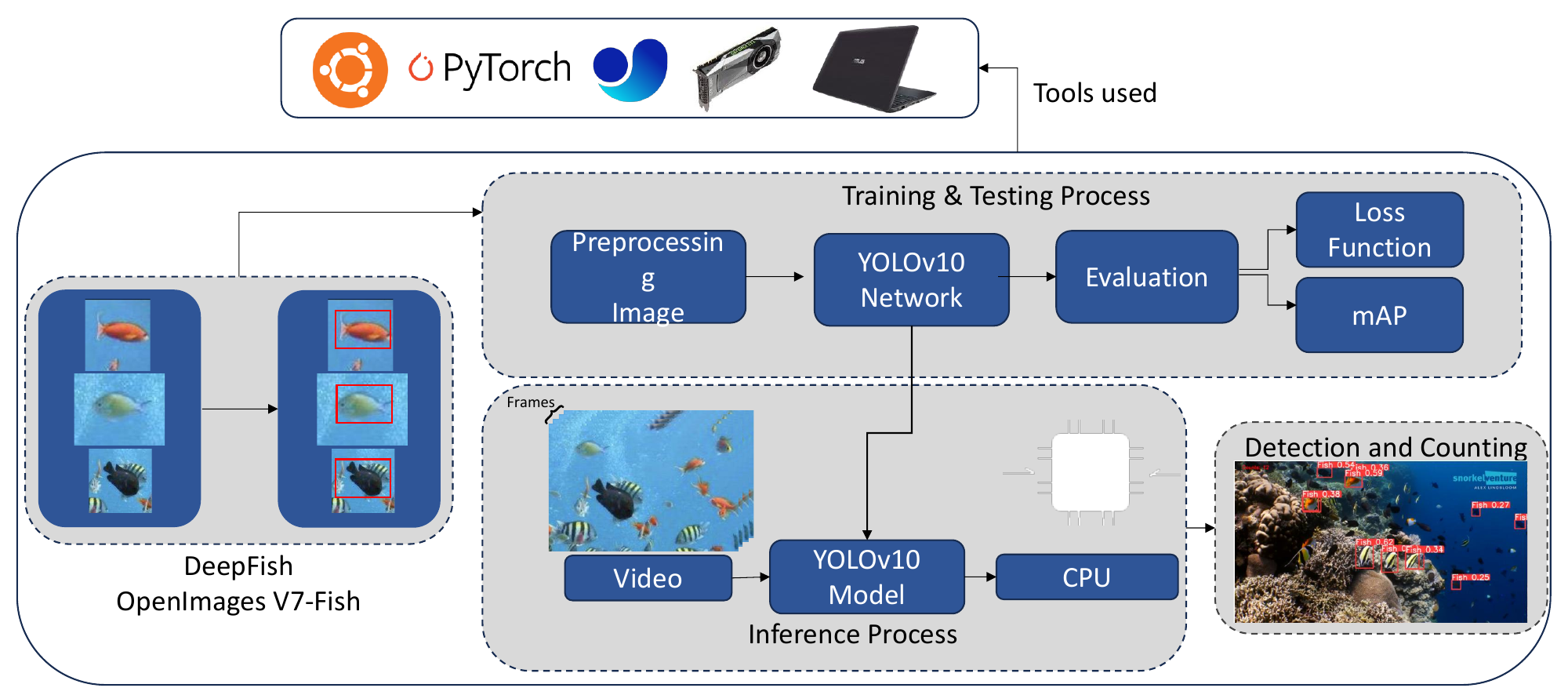}
    \vspace{-0.1in}
    \caption{System Design Concept
    }
    \label{fig: concept}
    \vspace{-0.15in}
\end{figure*}

\begin{table}[t!]
    \caption{Dataset Configuration}
    \centering
    \begin{tabular}{ccccc}
    \hline \hline
    Dataset & Train & Validation & Test & Total\\
    \hline
    Deep Fish & 3596 & 809 & 100 & 4505 images \\
    OpenImages V7-Fish & 5343 & 356 & - & 5699 images \\
    \hline \hline
    \end{tabular}
    \label{tab: dataset}
\end{table}

\begin{table}[t!]
    \caption{Training Configuration}
    \centering
    \begin{tabular}{p{1.5cm} p{2.5cm} p{2.5cm}}
    \hline \hline
    Parameters & Local Setup & Kaggle Setup\\
    \hline
    Platform & Ubuntu & Kaggle \\
    GPU & GTX 1080 Ti & Tesla P100 \\
    Image Size & $640 \times 640$ & $640 \times 640$ \\
    Epochs & 300 & 300 \\
    Batch Size & 32 & 32 \\
    Optimizer & Stochastic Gradient Descent (SGD) & Stochastic Gradient Descent (SGD) \\
    Learning Rate & 0.01 & 0.01 \\
    \hline \hline
    \end{tabular}
    \label{tab: training}
\end{table}

\section{Method}
\subsection{Research Procedure}
This research followed a systematic approach to effectively complete each task. The procedures involved:
\begin{enumerate}
    \item Information on YOLOv10, the object detection system used in the study, was gathered.
    \item Collected and prepared a fish detection dataset, including fish at the waterline and near-ocean floor levels.
    \item Set up the hardware and software environment required for training the detection model.
    \item Designed and developed the system and program to facilitate the model training process.
    \item Implemented the system and model in real-time testing environments for performance evaluation.
    \item Analyzed test results, focusing on accuracy, frames per second (FPS), and latency to assess the model's effectiveness.
\end{enumerate}

\begin{table}[h!]
    \caption{Inference Configuration}
    \centering
    \begin{tabular}{p{2.5cm} p{5cm}}
    \hline \hline
    Parameters & Local Setup \\
    \hline
    Platform & Ubuntu \\
    Compiler & Python 3.9.20 \\
    Framework & PyTorch 2.0 \\
    CPU & 12th Gen Intel(R) \hfill \break Core(TM) i7-12700 @ 2,10Ghz \\
    \hline \hline
    \end{tabular}
    \label{tab: inference}
\end{table}

\begin{table}[h!]
    \caption{Model Comparison with DeepFish Dataset}
    \centering
    \begin{tabular}{ccccc}
    \hline
    \hline
    Model & Parameters & GFLOPs & mAP50 & mAP50:95\\
    \hline
    YOLOv3 \cite{MUKSIT2022101847} & - & - & 0.960 & -\\
    YOLO-Fish1 \cite{MUKSIT2022101847} & - & - & 0.961 & -\\
    YOLO-Fish2 \cite{MUKSIT2022101847} & - & - & 0.957 & -\\
    AquaYOLO \cite{Vijayalakshmi2025} & 53.1M & - & 0.960 & 0.453\\
    YOLOv10-n & 2.7M & 8.4 & 0.966 & 0.606\\
    \hline
    \hline
    \end{tabular}
    \label{tab: comparison}
\end{table}

\begin{figure*}
    \vspace{-0.05in}
    \centering
    \includegraphics[width=0.83\linewidth]{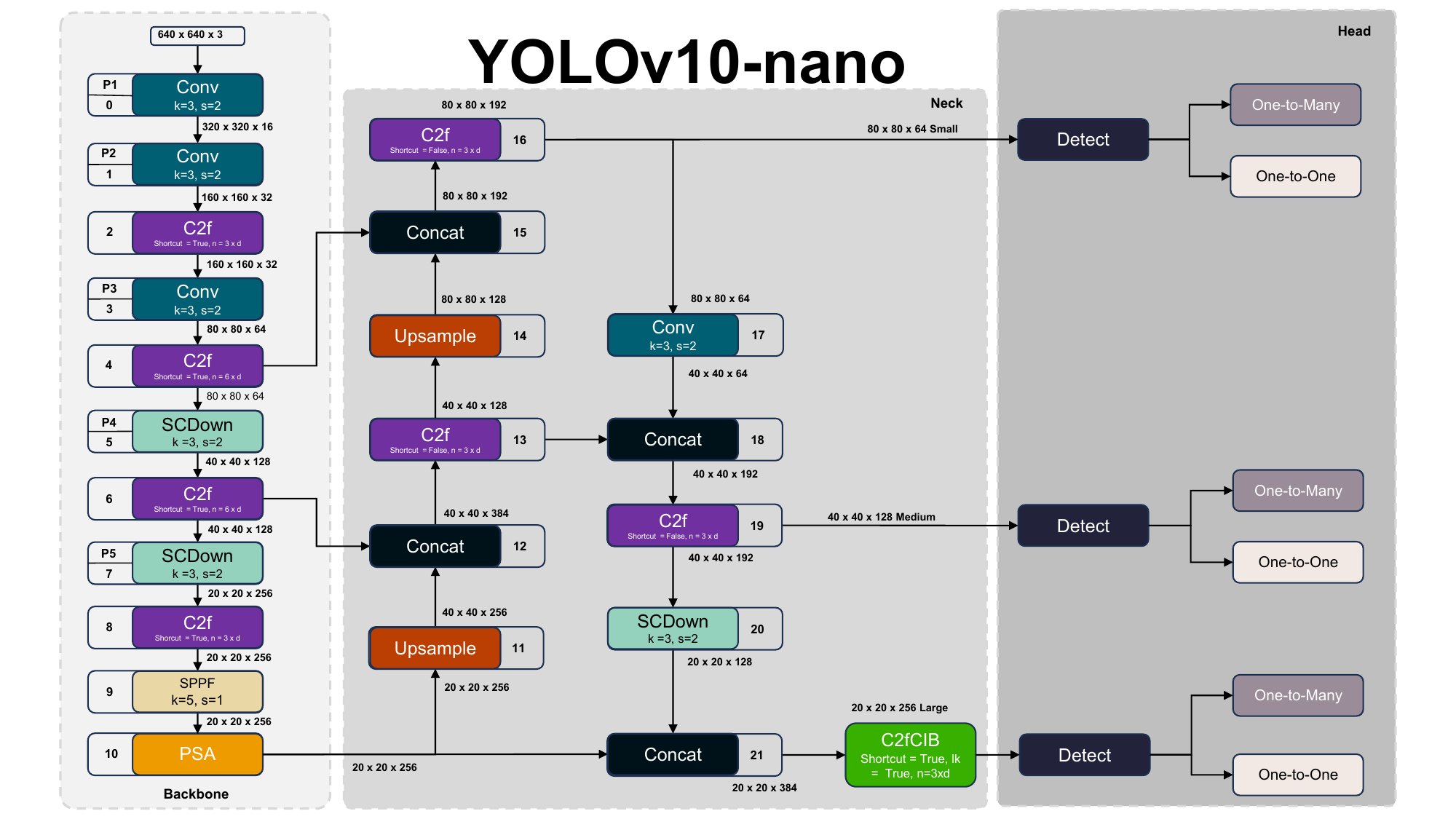}
    \vspace{-0.1in}
    \caption{YOLOv10-nano architechture
    }
    \label{fig: yolov10n}
    \vspace{-0.15in}
\end{figure*}

\subsection{System Design Concept}
This research focuses on building an object detection system and counting using YOLOv10 through a structured workflow. The core steps involve gathering and preprocessing fish detection datasets, then training the model to refine its parameters using a loss function. Finally, the model's performance is evaluated on new images or video frames, where it identifies objects by generating bounding boxes and labels. The entire system uses PyTorch, Ultralytics tools, and GPU acceleration for efficiency.

As shown in Figure \ref{fig: concept}, the development process begins with collecting the DeepFish and OpenImages V7-Fish datasets. After preparation, these images are pre-augmented from Ultralytics to ensure they're in the correct format for training. The YOLOv10 model is then trained on this combined dataset using a resolution of 640x640 pixels, a learning rate of 0.01, over 300 epochs, and a batch size of 32. This training includes tuning the loss function and adjusting model weights to enhance detection accuracy. Post-training, the model undergoes testing with GPU acceleration for speed and efficiency. Furthermore, its weights will be tested for inference using a CPU to confirm its capabilities in real-world scenarios.

\subsection{YOLOv10}
YOLO (You Only Look Once) v10, as shown in Figure \ref{fig: yolov10n}, launched in 2024, brought significant architectural improvements to the object detection framework. The model features an enhanced CSPNet backbone that improves feature extraction by optimizing gradient flow and eliminating unnecessary computations. For feature integration across different scales, YOLOv10 employs a Path Aggregation Network (PAN) \cite{miracle2020panet} in its neck section, enhancing its ability to detect objects regardless of their size. A significant innovation appears in YOLOv10's head section, which introduces dual prediction approaches: the One2One Head for inference, which generates single precise predictions per object and works best with well-separated targets; and the One2Many Head for training, which enables single anchor boxes to predict multiple objects in the same region—particularly valuable for detecting objects in crowded or overlapping scenarios \cite{Alzubaidi2021}, \cite{jmse13010135}.

Based on this new innovation in YOLOv10, this work chose YOLOv10, especially the nano variant, as its real-time detection, due to YOLOv10's performance and efficiency are superior to its predecessors (YOLOv5 and YOLOv8). While the predecessors of YOLO, such as YOLOv5 and YOLOv8, offer a good performance in detection, YOLOv10 achieves superior speed in inference with higher performance. The model's ability to balance performance and efficiency is vital for detecting and counting marine fish in their underwater environments \cite{isprs-archives-XLVI-3-W1-2022-301-2022}.

\subsection{Dataset}
Table \ref{tab: dataset} details the dataset configurations. The DeepFish dataset offers for detecting fish across 20 distinct mangrove habitats in Australia, to detect marine fish within Indonesian waters and to use as a benchmark dataset with other marine fish model such as YOLO-Fish1 and YOLO-Fish2, including "Rocky Mangrove - prop roots," "Sparse algal bed," and various others.

However, the DeepFish dataset proved insufficient for detecting fish near coral reef habitats in the same park, as it primarily focused on mangrove environments. To address this limitation, the model was also trained on the OpenImages V7-Fish dataset. This dataset contains more challenging fish images, such as fish tattoos, statues, and bones. Table \ref{tab: dataset} also outlines the training, validation, and test image configurations for the OpenImages V7-Fish dataset.

\subsection{Training Configuration}
Training for this research was conducted on two platforms to ensure efficiency and thoroughness. One environment was a local computer running Ubuntu Linux, equipped with an Intel Core i5-2320 CPU (3 GHz), 16 GB of RAM, and a GeForce GTX 1080 Ti GPU with 12 GB of memory. This setup facilitated 300 training epochs with an input resolution of 640x640 pixels, a batch size of 32, and a learning rate of 0.01. The Stochastic Gradient Descent (SGD) optimizer was chosen as its optimizer for its stability during this process.

In parallel, training also occurred on the Kaggle platform, leveraging an NVIDIA Tesla P100 GPU with identical configuration settings to accelerate the research. Table \ref{tab: training} shows the comprehensive setup for this phase in detail. Across both training environments, a loss function was consistently applied to measure the discrepancy between the model's predicted bounding boxes and object classes and the corresponding ground truth annotations. These configurations were uniformly applied to both the DeepFish and OpenImages V7-Fish datasets.

\subsection{Inference Configuration}
For the inference phase, the trained YOLOv10 model underwent evaluation on a local machine configured with a CPU. This setup was specifically selected to emulate real-world deployment scenarios where high-performance GPUs might not be accessible. Running inference on more constrained hardware is crucial for assessing the model's practical utility in field applications, such as underwater trash monitoring using low-power edge devices.

The inference environment was established on the Ubuntu operating system. It utilized Python 3.9.20 as the compiler and PyTorch 2.0 as the deep learning framework. The system was powered by a 12th Gen Intel® Core™ i7-12700 CPU, operating at 2.10 GHz. A comprehensive summary of the hardware and software configuration used during this inference stage can be found in Table \ref{tab: inference}.

%% file: 04_experiment.tex
\begin{figure}
    \vspace{-0.05in}
    \centering
    \includegraphics[width=0.83\linewidth]{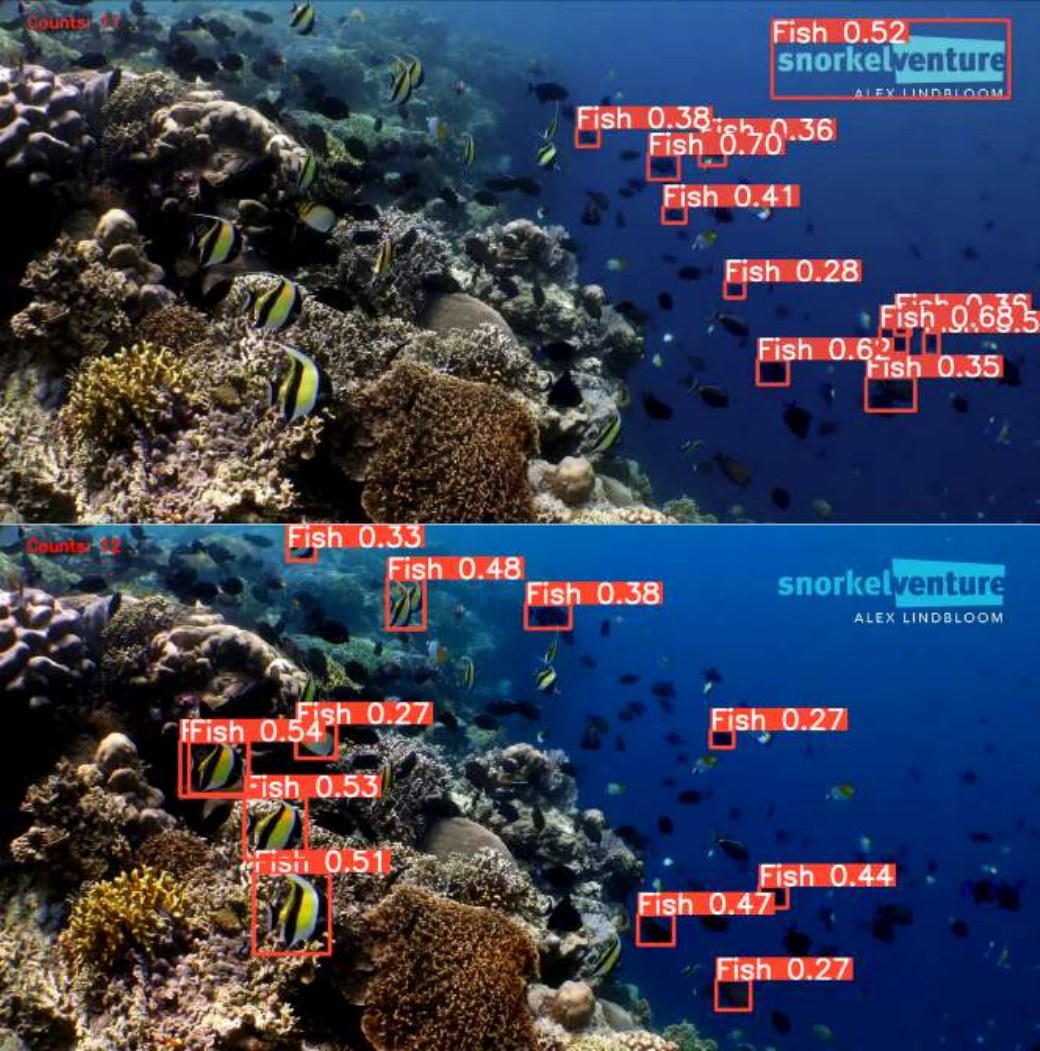}
    \vspace{-0.1in}
    \caption{Detection and counting comparison between the DeepFish (top) and OpenImages V7-Fish (bottom)
    }
    \label{fig: inference_comparison}
    \vspace{-0.15in}
\end{figure}

\begin{table}[t!]
    \caption{Model Performance with OpenImages V7-Fish Dataset}
    \centering
    \begin{tabular}{ccccc}
    \hline \hline
    Model & Parameters & GFLOPs & mAP50 & mAP50:95\\
    \hline
    YOLOv10-n & 2.7M & 8.4 & 0.505 & 0.344\\
    \hline \hline
    \end{tabular}
    \label{tab: comparison2}
\end{table}

\begin{table}[t!]
    \caption{Inference Speed}
    \centering
    \begin{tabular}{ccc p{1cm} p{1.3cm}}
    \hline \hline
    Model & Parameters & GFLOPs & Average FPS & Average Latency (s)\\
    \hline
    YOLOv10-n & 2.7M & 8.4 & 29.29 & 0.0324\\
    \hline \hline
    \end{tabular}
    \label{tab: inference_speed}
\end{table}

\section{Result and Discussion}
This section outlines the experimental results of the YOLOv10-nano model for marine fish detection. The discussion will cover the model's performance evaluation, a comparison with earlier work, an assessment of the dataset utilized, and an analysis of the model's inference speed. This aims to confirm the proposed approach's effectiveness and practical utility in real-world scenarios.

\subsection{Model Comparison}
Performance and computational efficiency differentiation with earlier works in marine fish detection is necessary to scientifically prove the robustness and efficiency of YOLOv10-nano in marine fish detection. This comparison is done by comparing the performance and computational needs of each model performed and use, respectively on a benchmark dataset, which is the DeepFish dataset. Table \ref{tab: comparison} presents the results in comparison with the YOLOv10-nano and earlier works.

Based on Table \ref{tab: comparison}, the performance of YOLOv10-n surpassed earlier works in marine fish detection. YOLOv10-n achieves mAP50 of 0.966 and mAP50:95 of 0.606. YOLOv10-n achieves this with (Floating-Point Operations per Second) FLOPs of 8.4G and parameters of 2.7 million. Based on these findings, YOLOv10-n is superior to the earlier works of marine fish detection in terms of performance.

\subsection{Dataset Assesment}
While the DeepFish dataset is crucial in benchmarking the performance of  YOLOv10-nano, the model that was trained on the DeepFish dataset is unable to detect clearly on marine fish close to the coral reefs, like, for example, the sea garden in Bunaken National Marine Park as shown in Figure \ref{fig: inference_comparison}, to amend this, the usage of OpenImages V7-Fish is added to the training list. Figure \ref{fig: inference_comparison} presents the detection of the DeepFish dataset and OpenImages V7-Fish in the sea garden of Bunaken National Marine Park.

Based on Figure \ref{fig: inference_comparison}, the DeepFish model misdetects the logo as a fish. On the other hand, the OpenImages V7-Fish detects fish normally without misdetecting a logo. Table \ref{tab: comparison2} shows the performance of mAP50 and mAP50:95 on the OpenImages V7-Fish. The performance of the OpenImages V7-Fish dataset on mAP50 and mAP50:95 is 0.505 and 0.344, respectively. As this dataset is part of a larger dataset, the OpenImages V7, currently, there's no research on using only this dataset alone with a real-time detection architecture.

\subsection{Inference Assesment}
Parameters and GFLOPs of YOLOv10-n are already listed in Tables \ref{tab: comparison} and \ref{tab: comparison2}. Furthermore, the use of real-time detection needs to be validated with its average FPS and average latencies on a video. As this model needs to be operated in a real-world scenario on a system that runs on a CPU, instead of a GPU, it can operate in CPU-based devices or low-cost devices. Table \ref{tab: inference_speed} illustrates the average FPS and average latencies on a video running in real-time.

Based on Table \ref{tab: inference_speed}, YOLOv10-n gains an average FPS of 29.29 and an average latency of 0.0324 seconds. Based on these findings, YOLOv10-n is sufficient enough in operating fluidly in CPU-based devices based on the inference configuration in Table \ref{tab: inference}.

%% file: 05_conclusion.tex
\section{Conclusion and Recommendation}
\subsection{Conclusion}
This study investigated the effectiveness and efficiency of the YOLOv10-nano model for marine fish detection, particularly in real-time applications on CPU-based systems. The evaluation comprised model comparison, dataset analysis, and inference performance measurement.

Experimental results demonstrated that YOLOv10-nano outperforms earlier models in marine fish detection tasks. It achieved a mAP50 of 0.966 and mAP50:95 of 0.606, indicating high detection accuracy. With a computational requirement of 8.4 GFLOPs and 2.7 million parameters, the model strikes an excellent balance between accuracy and efficiency.

In terms of dataset performance, the DeepFish dataset, while valuable, showed limitations in complex underwater scenes such as those near coral reefs. These limitations were mitigated by training the model of YOLOv10-n with the OpenImages V7-Fish dataset, which significantly improved object detection fidelity in natural marine environments. However, OpenImages V7-Fish alone yielded a lower mAP50 and mAP50:95 of 0.505 and 0.344, respectively, suggesting its usefulness lies in supplementing rather than replacing domain-specific datasets.

Furthermore, YOLOv10-nano achieved an average inference speed of 29.29 FPS and a latency of 0.0324 seconds on CPU-only devices. These results confirm that the model is highly suitable for real-time deployment in low-cost or resource-constrained systems, thus broadening the scope for practical applications such as underwater monitoring, biodiversity assessments, and marine conservation efforts.

\subsection{Recommendation}
Based on the results obtained, the following recommendations are proposed:
\begin{enumerate}
    \item Model Optimization for Edge Deployment: While YOLOv10-nano performs efficiently on CPUs, further compression techniques such as quantization and pruning may be explored to enhance performance on ultra-low-power edge devices.
    \item Generalization Evaluation: Extend testing to various marine habitats globally to ensure the generalization capability of the model, particularly in unfamiliar underwater conditions.
    \item Real-World Application Trials: Conduct pilot studies with marine researchers or marine park authorities  to assess the model's practical impact and usability in live monitoring tasks.
\end{enumerate}